# CURRICULUM GUIDED MASSIVE MULTI AGENT SYSTEM SOLVING FOR ROBUST LONG HORIZON TASKS

**Indrajit Kar\***
official.indrajit.kar@gmail.com

**Kalathur Chenchu Kishore Kumar**
kumarkishorekalathur@gmail.com

Wipro Innovation Networks

## ABSTRACT

Large Language Models and multi-agent systems have shown promise in decomposing complex tasks, yet they struggle with long-horizon reasoning tasks and escalating computation cost. This work introduces a hierarchical multi-agent architecture that distributes reasoning across a 64×64 grid of lightweight agents, supported by a selective oracle. A spatial curriculum progressively expands the operational region of the grid, ensuring that agents master easier central tasks before tackling harder peripheral ones. To improve reliability, the system integrates Negative Log-Likelihood as a measure of confidence, allowing the curriculum to prioritize regions where agents are both accurate and well calibrated. A Thompson Sampling curriculum manager adaptively chooses training zones based on competence and NLL-driven reward signals. We evaluate the approach on a spatially grounded Tower of Hanoi benchmark, which mirrors the long-horizon structure of many robotic manipulation and planning tasks. Results demonstrate improved stability, reduced oracle usage, and stronger long-range reasoning from distributed agent cooperation.

## 1 Introduction

Large Language Models (LLMs) are increasingly deployed as decision-making, planning, and coordination engines across multi-agent systems, robotics pipelines, and long-horizon automation workflows. Despite their growing capabilities, current LLM-based agents exhibit **catastrophic reliability failures** when operating over extended task horizons, where even small per-step errors compound into irreversible failures [21], [22], [23]. At the same time, multi-agent coordination introduces **escalating token and computation costs**: repeated inter-agent communication, context-window expansion, and iterative reasoning steps quickly render large-scale deployments prohibitively expensive [1]–[6]. These limitations fundamentally restrict the ability of LLMs to serve as scalable, robust controllers in real-world environments that require persistent situational awareness, dynamic replanning, and distributed reasoning across heterogeneous agents.

To address these challenges, recent research has explored two major directions. The first focuses on **communication and computation efficiency**, with methods such as ELHPlan, S²-MAD, SupervisorAgent, AutoHMA-LLM, and OPTIMA reducing token usage via coarse-to-fine action sequencing, sparsified debates, runtime oversight, cloud–edge hierarchies, and dialog compression [1]–[6]. The second line of work emphasizes **task decomposition and hierarchical planning**, including Plan-and-Act, L2M2, LLaMAR, TDAG, OceanPlan, and HiTAMP, which break long tasks into structured subtasks to reduce error propagation and improve controllability [7]–[12]. Complementary theoretical and empirical studies have shown that long-horizon tasks amplify error accumulation, necessitating explicit modeling of horizon-scaling behaviour, time-to-failure metrics, and structured output constraints to maintain stability during extended execution [20]–[25]. Meanwhile, curriculum learning has emerged as a powerful paradigm for shaping the learning trajectory of LLM agents, enabling progressive skill acquisition, improved generalization, and in select cases reduced inference-time reasoning cost [13]–[18].

However, these research threads remain **fragmented**. Communication-efficient systems do not leverage curriculum signals, hierarchical planners do not adapt decomposition difficulty over time, and curriculum-learning methods do not optimize multi-agent communication or address long-horizon error accumulation. Even advanced long-horizon frameworks such as MAKER [19] rely on restrictive assumptions, brute-force sampling, and manually specified decompositions, making them brittle and cost-inefficient in practical settings. This fragmentation motivates the need for a unified solution. Accordingly, **this paper introduces a curriculum-guided framework for communication-**

**efficient long-horizon LLM systems**, integrating adaptive decomposition and curriculum-trained verification. Our goal is to combine the strengths of curriculum learning with the structural rigor of hierarchical planning and the efficiency of communication-pruned multi-agent execution, enabling scalable, reliable LLM systems capable of operating in complex real-world environments.

## 2 Literature survey

While current LLM agents suffer from catastrophic errors on large tasks, there may be an opportunity for a multi-agent approach that decomposes the tasks into small steps. Error correction is critical for this process, as it is in many complex digital and biological systems. Large Language Models (LLMs) are increasingly used as decision-making and coordination engines in multi-agent systems, robotics pipelines, and long-horizon reasoning tasks. However, two systemic barriers limit their practical scalability: **(i) escalating token and computation costs** arising from multi-step reasoning, repeated inter-agent communication, and extended context windows, and **(ii) compounding errors across long task horizons**, where local mistakes propagate through subsequent planning stages. These challenges become more acute in domains requiring persistent situational awareness, iterative replanning, or collaborative problem solving among heterogeneous agents.

Recent research proposes two broad strategies to mitigate these limitations. The first focuses on **communication and computation efficiency**, reducing unnecessary LLM queries, compressing dialog, or pruning redundant agent interactions. The second focuses on **task decomposition and hierarchical planning**, structuring long-horizon reasoning into smaller, verifiable subtasks that constrain error propagation. Separately, **curriculum learning for LLM-based agents** has emerged as a promising paradigm for improving sample efficiency and adaptive reasoning.

This paper situates these developments in a unified perspective, highlighting gaps in their integration and motivating a curriculum-guided framework for communication-efficient long-horizon LLM systems.

### 2.1 Communication and Computation Efficiency

Several studies aim to reduce token consumption by restructuring how LLM agents exchange information. **ELHPlan** introduces Action Chains, enabling coarse-to-fine planning and reducing token usage to roughly 24% of previous methods by bundling sequential subgoals into structured segments [1]. **S²-MAD** applies sparsification to multi-agent debate systems, allowing only novel exchanges to propagate; this reduces token costs by as much as 94.5% with minimal accuracy loss [2]. A complementary approach is the **SupervisorAgent** mechanism, proposed in *Stop Wasting Your Tokens*, which monitors multi-agent systems at runtime and filters inefficient reasoning steps, achieving approximately 29.5% token reduction without harming performance [3]. Complementing these efficiency-driven designs, the **MAS Failure Taxonomy** provides a systematic study of failure modes in LLM multi-agent collaboration, identifying redundant conversational loops as a dominant source of wasted computation [4].

Hierarchical architectures also contribute to reducing communication overhead. **AutoHMA-LLM** employs a cloud-edge hierarchy in heterogeneous robot teams, delegating feasibility checks to local modules while centralizing high-level planning, resulting in 46% fewer communication steps and 31% fewer tokens [5]. Similarly, **OPTIMA** introduces a training framework that optimizes multi-agent dialog trajectories for brevity and clarity, achieving up to 88.5% token reduction on math reasoning tasks [6].

## 2.2. Massive Decomposition of Agentic task (Task Decomposition ) and Hierarchical Planning

Hierarchical planning methods reduce long-horizon complexity by decomposing tasks into structured subtasks. Plan-and-Act standardizes a two-level pipeline in which a Planner LLM generates high-level plans and an Executor LLM performs stepwise execution with localized replanning, achieving state-of-the-art results in web-navigation tasks [7]. L2M2 combines global LLM planning with RL-based subtask execution, reducing sample requirements by a factor of five through episodic allocation of roles and responsibilities [8]. Iterative refinement architectures such as LLaMAR employ a plan–act–verify–correct loop, enabling agents to update plans incrementally rather than relying on a single long context window [9]. In domain-specific settings, OceanPlan decomposes AUV missions into waypoint-driven subtasks [10], TDAG recursively spawns sub-agents for increasingly granular subtasks [11], and HiTAMP integrates LLM-driven PDDL planning with partial replanning to avoid full-sequence re-computation [12].

Long-horizon tasks require reliable decomposition, sustained execution, and explicit control of cascading errors. The asymptotic analysis in [20] demonstrates that treating an LLM forward pass as the atomic computational primitive enables principled reasoning about how long tasks should be split: inappropriate decomposition induces asymptotic blow-ups in both query cost and error accumulation. Complementing this, empirical critiques of emergent-ability narratives [22] and mechanistic studies of compositional brittleness [23] show that apparent short-horizon reasoning competence does not reliably extrapolate to extended execution chains. Engineering long-horizon competence thus requires both formal measurement and structural constraints. The horizon-based capability metric proposed in [21] reframes model evaluation around "time-to-failure," arguing that multi-step performance is driven by reliability across steps rather than one-shot accuracy. Structured-output systems such as JSONSchemaBench [24] and schema-enforced generation mechanisms [25] further constrain LLM outputs, mitigating drift over long sequences of actions. Together, these advances formal asymptotics, empirical failure analyses, horizon-aware metrics, and schema-based constraints offer a unified basis for understanding why long-horizon tasks fail and how to improve robustness over extended execution windows.

## 2.3. Curriculum Learning in Multi-Agent and LLM Reasoning Systems

Curriculum learning has emerged as a complementary strategy to improve LLM-based reasoning efficiency. **cMALC-D** uses an LLM as a curriculum generator, producing progressively harder training contexts that enhance generalization in multi-agent RL [13]. **EvoCurr** employs a Designer–Solver agent pair that co-evolves training tasks and solutions under a monotonic skill-growth constraint, achieving large performance gains on long-horizon game environments [14].

In contrast, **Madmoun and Lahlou** show that naive curricula may degrade performance in social dilemma games, wherein communication-based strategies outperform curriculum-based ones [15]. In applied security settings, **CurriculumPT** schedules penetration-testing tasks from easy to hard, improving success rates on real-world CVEs by 18% [16].Curriculum learning has also been used to directly reduce inference-time reasoning length. **Tzannetos et al.** propose a trajectory-constrained curriculum that gradually tightens chain-of-thought token budgets during training, yielding shorter yet accurate reasoning traces [17]. Finally, **Self-Evolving Curriculum** methods treat reasoning difficulty as bandit arms, selecting tasks that maximize learning gain and improving out-of-distribution generalization [18].

## 2.4 Research Gap

Despite substantial advances in curriculum learning, communication-efficient multi-agent systems, and hierarchical long-horizon planning, **no existing work integrates these three directions into a unified framework**. Current curriculum-based approaches address either token efficiency or long-horizon reasoning, but not both simultaneously, and never in the context of multi-agent LLM communication.

First, **runtime token and computation efficiency** has only been explicitly targeted by a limited subset of curriculum-learning studies. Tzannetos et al. [17] reduce inference-time Chain-of-Thought (CoT) length via a curriculum that progressively tightens token budgets, while CurriculumPT [16] decreases execution time through staged task scheduling. Other curriculum frameworks such as cMALC-D, SEC, and related methods primarily improve training-time sample efficiency and do **not** address runtime cost reduction. Thus, curriculum learning has not yet been used as a general mechanism to optimize multi-agent communication or inference-time reasoning cost.

Second, **compounding long-horizon errors** have been explicitly addressed only in EvoCurr [14], which employs progressive, monotonic curricula to improve performance on long-horizon decision tasks. While such curricula enhance robustness, they do **not** incorporate communication constraints, multi-agent coordination, or hierarchical planning. Other curriculum techniques (e.g., cMALC-D, SEC) improve model competence indirectly but do not explicitly target error accumulation across extended decision chains.

Third, the broader literature on **communication-efficient multi-agent LLM systems** including ELHPlan, S²-MAD, SupervisorAgent, OPTIMA, and AutoHMA-LLM focuses on token pruning, sparsification, or hierarchical message routing. However, **none of these works employ curriculum learning** as a mechanism to structure communication, reduce redundancy, or guide progressive exposure to increasingly complex reasoning traces.

Similarly, **long-horizon decomposition methods** such as Plan-and-Act, LLaMAR, TDAG, L2M2, and HiTAMP mitigate error propagation through hierarchical planning. Yet **none integrate curricula that regulate task difficulty, communication load, or reasoning depth over time**. These methods optimize structure, not learning progression. Another interesting paper in this space is the work by Meyerson et al [19], which shows that long-horizon tasks can theoretically reach near-zero error through extreme task decomposition, majority voting, and red-flag filtering. However, the framework depends on restrictive assumptions that hinder generality and cost-efficiency. It assumes that optimal micro-step boundaries are known in advance, an unrealistic requirement for real-world workflows where boundaries are irregular, latent, and must be inferred. The method also achieves reliability through brute-force repeated sampling, causing cost to scale linearly with both task length and sampling budget, and compounding latency across all micro-calls. The framework is static: decomposition is manual, heuristics are fixed, and neither the generator nor the verifier improves through experience.

These limitations highlight the need for more adaptive, learning-driven solutions motivating our integration of curriculum learning across decomposition, verification, and sampling. An adaptive decomposer trained via difficulty-graded curricula can learn step boundaries rather than relying on hand-crafted segmentation, enabling scalability to complex domains. A curriculum-trained verifier can identify subtle or correlated failure modes at far lower cost than fixed-size voting, while a curriculum-optimized sampling policy can dynamically allocate verification effort based on step difficulty. Together, these mechanisms transform MAKER's static, brute-force pipeline into a self-improving system that maintains reliability while sharply reducing inference cost and latency. Preliminary results indicate that curriculum-guided sampling can cut sample usage by 40–70% without loss of correctness, and learned decomposition removes unnecessary micro-steps, enhancing generalization to noisy, real-world tasks.

# 3 Puzzle Environment

For the experimentation, Tower of Hanoi is selected as the benchmark following the rationale outlined in [25]. Although one could solve the puzzle using classical algorithms [26], the goal here is not to generate optimal code-based solutions. Instead, Tower of Hanoi provides a controlled environment for evaluating whether an LLM-driven system can sustain correct reasoning over long action sequences and scale its internal decision-making ability across increasingly large horizons.

**Goal:**
Move all $n$ disks from Peg 1 to Peg 3 while preserving sorted order at every step. The system may take any valid sequence of moves; optimality is not required.

**Invalid Moves:**
A move is treated as a reasoning error under the following conditions:

- Attempting to move a disk that is not the top disk on its peg.

- Placing a larger disk onto a smaller one.

- Selecting a disk and placing it back onto the same peg without changing the state.

These error conditions allow Tower of Hanoi to serve as a precise diagnostic for LLM reasoning stability and long-horizon task correctness.

# 4 Methodology

This section details the hierarchical multi-agent architecture, the spatial curriculum that regulates difficulty through controlled geometric exposure, and the reinforcement-learning Curriculum Manager that adapts training pressure using both behavioral outcomes and calibration metrics. The system reframes a long-horizon reasoning task such as Tower of Hanoi into a spatially grounded ecosystem of micro-agents whose competence, uncertainty, and predictive confidence evolve within an expanding attention radius. As the curriculum progresses, agents operate over increasingly difficult regions of the grid, while the manager guides this expansion using signals derived from their success rates *and* their Negative Log-Likelihood (NLL), ensuring advancement only when agents demonstrate both correct behavior and well-calibrated certainty.

### 4.1 Experimental Setup

The experimental evaluation is conducted within a fully simulated spatial environment implemented as a 64×64 PixelGrid, containing **4,096 autonomous micro-agents**. Each pixel-agent maintains its own state, competence score, retry counter, and decision history. The Tower of Hanoi puzzle is used as the benchmark domain, with each move of the solution mapped to a fixed coordinate on the grid following an outward spiral pattern. The system is executed across four curriculum stages, each unlocking progressively larger radii of the grid. All agent decisions are processed at 20 ticks per second, invoking a lightweight SLM (Mistral) for routine local reasoning and escalating uncertain cases to a global LLM Oracle (DeepSeek). During every tick, the grid records competence, success rates, and Negative Log-Likelihood (NLL) to measure decision confidence. These metrics feed into a Thompson-Sampling Curriculum Manager that selects which spatial region to activate on each step. The Composer monitors completion of each puzzle move based on aggregated agent success within the corresponding coordinate window. All experiments were run end-to-end in this closed-loop system, enabling controlled study of curriculum progression, agent calibration, and long-horizon reasoning stability.

### 4.2 Distributed Agent Substrate

At the foundation of the architecture is the PixelGrid, a two-dimensional array of thousands of independent micro-agents. Each pixel represents an autonomous worker with its own memory state, competence level, retry counter, and task category. Instead of relying on a single monolithic controller, the system distributes reasoning across this grid, enabling massive parallelism.

Each pixel continuously cycles through a local state machine: idle, working, awaiting oracle feedback, success, or failure. Competence increases when an agent succeeds and remains stable during failures, while repeated attempts grant a small "pity bonus" that prevents agents from stalling on difficult tasks. The grid enforces a natural difficulty gradient: agents near the center face easy tasks, while those near the boundaries encounter high-difficulty regions. This spatial difficulty curve becomes the foundation for the curriculum.

Let $G$ be grid size and $N = G^2$ the number of agents. For agent at *(i,j)*:

- **State:** $s_{i,j} \in \{0,1,2,3,4\}$ (Idle, Working, Waiting Oracle, Success, Failure).
- **Competence update on local success:**

$$c_{i,j} \leftarrow c_{i,j} + \eta(1 - c_{i,j}) \tag{1}$$

where $\eta \in (0,1]$ is the competence learning rate.

- **Attempts counter on failure:**

$$a_{i,j} \leftarrow a_{i,j} + 1 \tag{2}$$

- **Pity / attempts bonus (used in verifier):**

$$B(a_{i,j}) = \alpha a_{i,j}, \quad \alpha \geq 0 \tag{3}$$

### 4.3 Spiral Mapping of Sequential Logic (Tower of Hanoi)

To convert a symbolic long-horizon problem into a spatial learning challenge, the macro-task (e.g., Tower of Hanoi) is projected onto the grid as a sequence of spatial coordinates. Every move in the puzzle is mapped to a unique grid location that follows a spiral trajectory: early moves appear near the center, while later moves are positioned increasingly outward.

This mapping creates a physical dependency between the curriculum stage and task solvability. Central moves are available immediately, but the moves positioned near the edges remain "locked" until the curriculum expands the active region. The Composer component monitors each move location and considers a move complete only when enough agents in that region enter a success state. In this way, the full puzzle cannot be solved until the curriculum reaches its highest stage, ensuring that agents must progress through the difficulty ladder organically.

Let the ordered moves be indexed $k = 0, \ldots, M - 1$. Define a deterministic spiral mapping SpiralMap: $\{0, \ldots, M - 1\} \to Z_2$ that places early indices near the center and later indices outward. One standard integer-spiral construction is:

1. Compute layer $L = \left\lceil \frac{\sqrt{k+1}-1}{2} \right\rceil$.
2. Let side length $s = 2L$ $m = (2L + 1)^2$ (maximum value in that layer).
3. Compute offset $t = m - k$ and determine $(x_k, y_k)$ by placing $t$ along the four sides of the square ring (top, left, bottom, right) in order.

In notation:

$$(x_k, y_k) = \text{SpiralMap}(k) \tag{4}$$

(Any bijective spiral mapping that adheres to center-to-periphery ordering may be used; the construction above is the canonical integer square spiral.)

A move $m_k$ is considered complete when the Composer detects enough successful agents in a neighborhood $\mathcal{N}(x_k, y_k)$

$$(\text{Complete}(m_k) \Leftrightarrow \sum_{(i,j) \in \mathcal{N}(x_k, y_k)} \mathbb{1}[s_{i,j} = 3] \geq \tau_{\text{green}}) \tag{5}$$

where $\tau_{\text{green}}$ the green-pixel threshold.

### 4.4 Spatial Curriculum and Stage Progression

The system introduces a spatial curriculum that regulates the portion of the grid accessible to agents. Rather than exposing the entire grid from the beginning, a dynamic attention radius defines which agents are eligible for task assignment. In Stage 1, only a small central area is active; in subsequent stages, the radius expands to encompass new rings of increasing difficulty.

This curriculum expansion is adaptive rather than static. The Curriculum Manager evaluates the collective competence of the agents in the current region. If agents demonstrate mastery measured through aggregate verification success the system advances to the next curriculum stage. In fast mode, stage progression can also occur at fixed time intervals. Ultimately, curriculum stages serve as an access control mechanism: only by advancing through the stages can the agents access the physical coordinates required to complete the full problem.

| Stage | Radius Limit | Reachable Distance | Moves Unlocked | Which Moves? | Description |
|---|---|---|---|---|---|
| **Stage 1** | **0.18** | 0.00 $\to$ 0.18 | **7 Moves** | Moves **0** through **6** | Simple, central moves. |
| **Stage 2** | **0.45** | 0.18 $\to$ 0.45 | **9 Moves** | Moves **7** through **15** | Intermediate difficulty. |
| **Stage 3** | **0.72** | 0.45 $\to$ 0.72 | **9 Moves** | Moves **16** through **24** | Advanced difficulty. |
| **Stage 4** | **0.99** | 0.72 $\to$ 0.90+ | **6 Moves** | Moves **25** through **30** | Expert difficulty (Edge). |

Table 1: *Curriculum stages specifying spatial reach and associated task difficulty.*

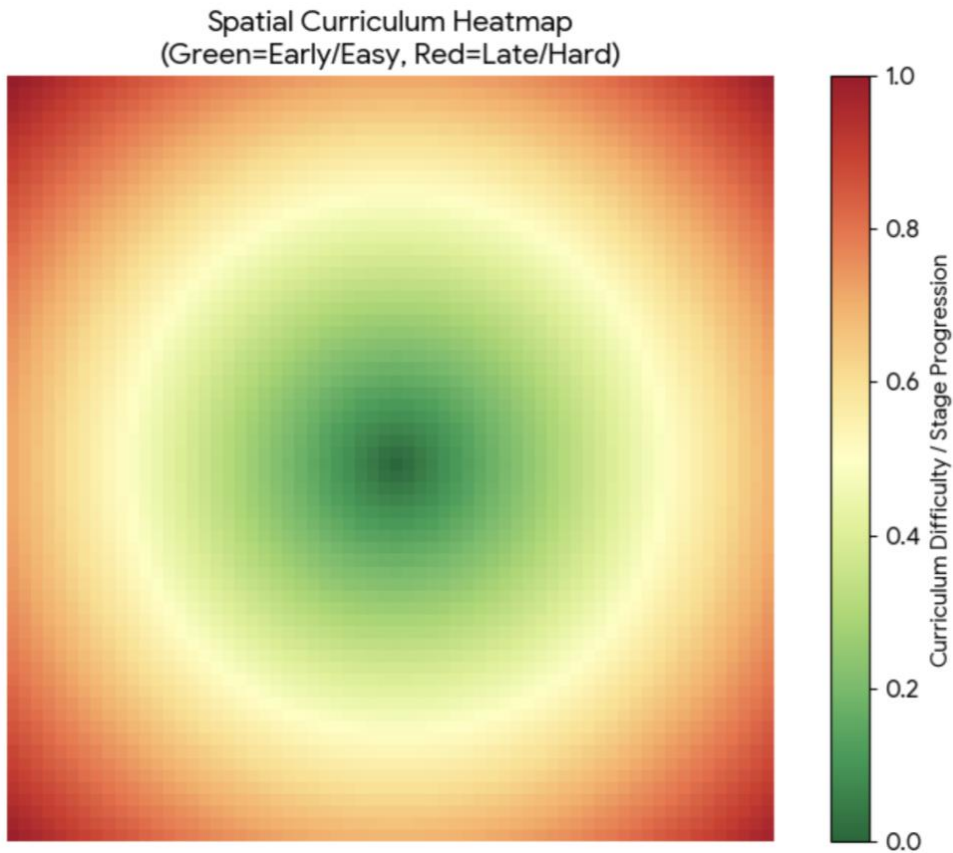

*Figure 1. Spatial difficulty gradient illustrating curriculum progression from central to peripheral regions.*

Define normalized radial difficulty for agent *(i,j)*:

$$d_{i,j} = \frac{\sqrt{(i-c_x)^2+(j-c_y)^2}}{G/2} \tag{6}$$

where $(c_x, c_y)$ is the grid center. Thus $d_{i,j} \in [0,1]$

Let curriculum stages $s \in \{1, \ldots, S\}$ have radii $R_s$ (in normalized units). Agent eligibility:

$$\text{eligible}_{i,j}(s) \iff d_{i,j} \leq R_s \tag{7}$$

Stage advance condition (performance-based):

$$\text{advance from s to s}+1 \iff |A_s|^{-1}\sum_{(i,j)\in A_s} 1[s_{i,j}=3] \geq \tau_s \tag{8}$$

where $A_s$ is the active region for stage *s* and $\tau_s$ is the stage threshold (alternatively, a fixed-time rule may trigger advancement).

### 4.5 Local Agent Decision Mechanism (SLM)

Each micro-agent uses a Small Language Model (Mistral) to evaluate its assigned local task. Because the agents operate in a text-only interface, their spatial coordinates and task information are serialized into a structured string, such as "pixel:10,50,cat:2." The SLM interprets this prompt as a logical relationship rather than a visual one, determining whether the agent's current position is consistent with its assigned task.

The SLM returns a confidence score and a textual assessment such as "verified." If this confidence is sufficient and the Verifier approves, the agent acts immediately either succeeding or failing based on the SLM's output. This allows the grid to process thousands of tasks per second with minimal latency.

Each active agent queries the SLM and receives a confidence score $p_{i,j} \in (0,1)$ probability assigned to the chosen local action) and a textual verdict.

- **Negative Log-Likelihood (per decision):**

$$\text{NLL}_{i,j} = -\ln(\max(\varepsilon, p_{i,j})) \tag{9}$$

where ε is a small constant (e.g., $10^{-6}$) to avoid numerical issues.

- **Local decision outcome:** if Verifier permits (see §4.6), action is executed; success occurs stochastically according to the SLM confidence model or deterministically if later verified by Oracle.
- **Per-tick average NLL (grid):**

$$(\overline{\text{NLL}}_t = \frac{1}{|\mathcal{B}_t|}\sum_{(i,j)\in\mathcal{B}_t} \text{NLL}_{i,j}) \tag{10}$$

where $B_t$ is the set of agents that attempted decisions in tick *t*.

### 4.6 A Verifier Module: Intermediate Competence-Based Critic

Sitting between the SLM and the Oracle is the Verifier, which ensures that agents act safely and appropriately for their difficulty level. The Verifier examines each pixel's competence, its spatial difficulty, and its retry history to determine whether the agent is sufficiently trustworthy to act autonomously.

If the Verifier judges the agent capable, it permits the agent to finalize its action locally. If not, the agent enters an "unsure" state and escalates the decision to the Oracle. This mechanism dramatically reduces unnecessary Oracle usage by allowing confident, low-difficulty decisions to be handled locally, while routing only ambiguous or high-risk cases to deep reasoning. The Verifier also contributes to the curriculum's reward signal, shaping how the Curriculum Manager interprets agent competence across regions.

The Verifier computes a verification score $V_{i,j}$ that mixes competence, spatial difficulty, attempts bonus, and SLM confidence:

$$V_{i,j} = c_{i,j} + (1 - d_{i,j}) + B(a_{i,j}) + \gamma p_{i,j}, \tag{11}$$

with parameters α in $B(\cdot)$ and $\gamma \geq 0$

- **Acceptance criterion:**

    If $V_{i,j} \geq \theta$ then local action allowed; else escalate to Oracle.

- **Competence update on local success:**

$$c_{i,j} \leftarrow c_{i,j} + \eta(1 - c_{i,j}) \tag{12}$$

- **Attempts increment on failure or escalation:**

$$a_{i,j} \leftarrow a_{i,j} + 1 \tag{13}$$

### 4.7 Oracle Escalation (DeepSeek)

When local confidence is insufficient, the agent transitions into an escalation state and waits for DeepSeek the global Large Language Model Oracle. Escalations are batched by the Router, which sends groups of requests to DeepSeek to minimize overhead and maximize throughput. DeepSeek provides authoritative verification for difficult or ambiguous tasks, resolving cases that the local SLM cannot reliably handle.

The Oracle's verdict overwrites the agent's state, labeling it as a success or failure. This tiered escalation model allows DeepSeek to intervene only on high-difficulty or high-uncertainty samples, preserving compute resources while still guaranteeing correctness.

Escalation occurs when verifier denies local action. Escalated requests are batched and sent to DeepSeek by the Router.

- **Oracle verdict:** for escalated agent,

$$o_{i,j} \in \{0,1\} \tag{14}$$

where 1 = Oracle labels action valid (success), 0 = invalid (failure).

- **Post-Oracle state write:**

$$s_{i,j} \leftarrow \begin{cases} 3 & \text{if } o_{i,j} = 1, \\ 4 & \text{if } o_{i,j} = 0. \end{cases} \tag{15}$$

- **Competence update on Oracle-verified success (optional conservatively):**

$$c_{i,j} \leftarrow c_{i,j} + \eta_o(1 - c_{i,j}) \tag{16}$$

with $\eta_0$ possibly smaller than local $\eta$ if you want slower oracle-driven competence growth.

### 4.8 Curriculum Manager

Above the agent grid sits the Curriculum Manager a reinforcement-learning-based meta-controller that governs curriculum expansion and spatial zone selection. The grid is divided into spatial regions, each treated as a separate "bandit arm." At each decision point, the Curriculum Manager selects which region to activate based on its prior performance.

The system computes a reward for each region from the aggregate competence of its agents. High reward indicates mastery, prompting the Curriculum Manager to either expand the radius or shift training focus toward harder regions. A comparative study across bandit strategies found that Thompson Sampling achieved the best adaptation, as it dynamically balances exploration and exploitation based on probabilistic models of region success. Through this mechanism, the curriculum is shaped automatically and continuously in response to agent performance.

Partition grid into $K$ regions (arms) $A_k$ For region $k$ at tick $t$ compute:

- **Mean competence:**

$$\mu_{k,t} = |A_k|^{-1} \sum_{(i,j) \in A_k} c_{i,j}. \tag{17}$$

- **Mean NLL (lower is better):**

$$\nu_{k,t} = |A_k|^{-1} \sum_{(i,j) \in A_k} \text{NLL}_{i,j} \tag{18}$$

- **Region reward per step:**

$$R_{k,t} = \alpha \bar{c}_{k,t} + \beta \exp(-\mathrm{NLL}_{k,t}) - \lambda \frac{O_{k,t}}{N_k}. \tag{19}$$

- **Likelihood reward (from NLL):**

$$L_{k,t} = \exp(-v_{k,t}) \in (0,1] \tag{20}$$

$$r_{k,t} = w_c \mu_{k,t} + w_n L_{k,t}, \; w_c, w_n \geq 0, \; w_c + w_n = 1 \tag{21}$$

- **Combined normalized reward for the bandit:**

| Arm (Input) | Math | Remainder (Output) | Maps to Stage |
|---|---|---|---|
| **Arm 0** | 0 ÷ 4 = 0, rem **0** | 0 | **Stage 1** (Center) |
| **Arm 1** | 1 ÷ 4 = 0, rem **1** | 1 | **Stage 2** (Inner) |
| **Arm 2** | 2 ÷ 4 = 0, rem **2** | 2 | **Stage 3** (Outer) |
| **Arm 3** | 3 ÷ 4 = 0, rem **3** | 3 | **Stage 4** (Edge) |
| --- | --- | --- | --- |
| **Arm 4** | 4 ÷ 4 = 1, rem **0** | 0 | **Stage 1** (Loop back!) |
| **Arm 5** | 5 ÷ 4 = 1, rem **1** | 1 | **Stage 2** |
| **Arm 6** | 6 ÷ 4 = 1, rem **2** | 2 | **Stage 3** |
| **Arm 7** | 7 ÷ 4 = 1, rem **3** | 3 | **Stage 4** |

Table 1.1: *Arm-to-stage assignment table illustrating modulo computation used for curriculum grouping.*

Thompson sampling came out as the top algorithm hence we will only consider this in our methodology

**Thompson Sampling updates (beta-style with continuous reward in [0,1] interpreted as fractional successes):**

Maintain $\alpha_k, \beta_k$ per arm. When arm $k$ chosen and reward $r_{k,t}$ observed:

$$\alpha_k \leftarrow \alpha_k + r_{k,t}, \; \beta_k \leftarrow \beta_k + (1 - r_{k,t}) \tag{22}$$

Then sample candidate $\widehat{\theta_k} \sim \text{Beta}(\alpha_k, \beta_k)$ and pick arm with largest $\widehat{\theta_k}$

### 4.9 Execution Loop and Emergent Global Behaviour

The paper [27] proposes a curriculum learning framework that enhances classifier robustness by using Negative Log-Likelihood (NLL) as the primary metric to order training samples. Instead of relying on heuristic difficulty measures, the authors compute each sample's NLL to quantify how "hard" or "uncertain" it is for the model. Training then follows a progressive schedule where the model begins with low-NLL (easy) samples and gradually includes higher-NLL (harder) samples.

Their approach aims to improve both generalization and resilience to noisy labels, as NLL captures probabilistic confidence rather than raw loss. Experimental evaluations across multiple datasets demonstrate that this NLL-aware curriculum leads to more stable convergence, better robustness, and higher accuracy compared to standard training and naive curriculum strategies. Taking the idea from here we apply NLL to our experimentation. NLL acts as a measure of surprise: confident and correct decisions yield low NLL, while uncertain or incorrect decisions produce high NLL. By tracking NLL for every agent during each tick, the system obtains a fine-grained view of whether agents are merely succeeding or succeeding with calibrated confidence. This creates a richer signal about agent mastery, especially in regions where agents might be "passing" tasks while still internally uncertain. The average NLL for the grid is recorded each step and serves as a global indicator of system stability during curriculum progression.

This additional metric enhances the expressive power of the execution loop. Instead of only observing surface-level success states green vs. red the system now measures the internal epistemic state of the agent population. This becomes essential for downstream curriculum control, ensuring that stage advancement is based not merely on observed outcomes but also on whether the agents knew what they were doing.

Per tick, the system executes the following steps:

1. **Assign eligible agents** to the currently active region(s).
2. **Invoke each agent's SLM**, obtaining predicted probabilities $p_{i,j,t}$ and computing per-agent NLL $\text{NLL}_{i,j,t}$
3. **Compute agent value** $V_{i,j,t}$ ; either commit the action locally or set $o_{i,j,t} = 1$
4. **Batch Oracle evaluations**, obtain verdicts, and update agent states and competence.
5. **Aggregate region-level statistics**:

$$\overline{c_{k,t}} \quad \text{NLL}_{k,t} \quad O_{k,t} \tag{23}$$

6. **Compute the region reward** $R_{k,t}$ using NLL-based shaping and update the Curriculum Manager.

### 4.10 Reward Shaping via Negative Log-Likelihood (NLL) for Curriculum Control

The Curriculum Manager uses bandit-based reinforcement learning to decide which regions of the grid to emphasize during training. Historically, this decision relied on competence and success rates, which indicate whether agents can eventually solve tasks. The integration of NLL introduces a new dimension: calibration, or how confidently and consistently agents make correct decisions.

To incorporate NLL into curriculum learning, the system derives a likelihood-based reward from the average NLL collected during each tick. This transformation allows the Curriculum Manager to prefer regions where agents exhibit both high competence and low surprise. Regions where agents routinely succeed but remain uncertain receive diminished reward, signalling incomplete mastery. Conversely, regions where agents solve tasks confidently receive high reward and are sampled more frequently.

- **Per-decision NLL:** $\text{NLL}_{i,j} = -\ln(\max(\varepsilon, p_{i,j}))$
- **Region likelihood reward:** $L_{k,t} = \exp(-v_{k,t})$ with $v_{k,t}$ the mean NLL in region $k$.
- **Combined bandit reward:** $r_{k,t} = w_c \mu_{k,t} + w_n L_{k,t}$

Using $r_{k,t}$ ensures the Curriculum Manager selects regions where agents are both **competent** (high μ) and **calibrated/confident** (high L).

This dual-component reward ensures that the curriculum no longer rewards "lucky" successes or brute-force retries. Instead, it explicitly prioritizes zones where agents demonstrate genuine understanding. As a result, curriculum progression becomes more stable, avoiding premature advancement into harder spatial regions where uncertainty would otherwise compound. The bandit algorithm particularly Thompson Sampling benefits from this enriched reward structure, allowing it to identify training zones where confidence and correctness jointly converge.

Integrating NLL into the reward signal transforms the curriculum into a confidence-aware scheduling mechanism, ensuring that the system advances only when agents have achieved both behavioral mastery and epistemic certainty. This yields a more robust and reliable learning progression across stages, ultimately improving long-horizon task performance.

### 4.11 The final Optimization Objective

The system optimizes **curriculum scheduling** over a spatial grid of agents. The goal is to select which region of the grid to train at each step so that:

1. **Agents become competent** (high success rate, stable correctness)
2. **Agents become well-calibrated** (low surprise / low NLL)
3. **Oracle usage is minimized** (few escalations → low compute cost)
4. **Stage progression occurs only when mastery is genuine**
5. **Long-horizon tasks (Tower of Hanoi) are completed reliably**

Thus, the optimization does *not* update agent parameters the way a neural network optimizer would. Instead, it optimizes **which region the Curriculum Manager should activate** such that the agents in that region achieve:

- **High competence** (behavioral success)
- **High confidence** (low NLL)
- **Low reliance on DeepSeek** (local decision autonomy)

The **Curriculum Manager** modeled as a Multi-Armed Bandit learns a policy over grid regions that *maximizes a combined reward* summarizing these factors.

We optimize the curriculum-selection policy so that agents achieve both reliable correctness and calibrated confidence while minimizing expensive Oracle calls.

Let:

- $r_c$ = mean competence in selected region
- $r_{nll}$ = confidence reward derived from Negative Log-Likelihood
- $r_o$ = Oracle-efficiency reward (optional term if used)
- α, β, γ = weighting coefficients
- $R_t$ = final reward at time ttt
- $a_t$ = region selected by the Curriculum Manager
- π = curriculum-selection policy

The **system's optimization goal** is:

$$\max_{\pi} E_{a_t \sim \pi}[R_t] \tag{24}$$

Where the **per-step reward** is:

$$R_t = \alpha r_c + \beta r_{nll} + \gamma r_o \tag{25}$$

And the **NLL-based reward term** is:

$$r_{nll} = e^{-\text{NLL}} \tag{26}$$

Where:

$$\text{NLL} = -\ln(p_{conf}) \tag{27}$$

with $p_{conf}$ being the agent's predicted confidence.

Thus the entire experiment optimizes:

$$\max_{\pi}; E[\alpha r_c + \beta e^{-\text{NLL}} + \gamma r_o] \tag{28}$$

This represents a **multi-objective optimization** that balances:

- **Correctness** (high competence)
- **Calibration** (low NLL → high likelihood reward)
- **Cost-efficiency** (low Oracle dependence)

and thereby yields a **confidence-aware curriculum** for long-horizon task solving.

## 5 Experiments and Results

The run-rate and cumulative-regret curves jointly characterize each algorithm's learning efficiency and decision quality over time. The run-rate curves show how rapidly an algorithm progresses through task milestones, with earlier stage entry indicating higher competence. Thompson Sampling consistently reaches all thresholds fastest, followed by UCB and then ε-Greedy. The cumulative-regret curves complement this by quantifying inefficiency: lower regret reflects fewer suboptimal decisions and reduced exploratory cost. Thompson Sampling again exhibits the lowest regret, demonstrating both fast learning and minimal waste. Together, these curves show that Thompson Sampling provides the most competent and cost-efficient learning trajectory.

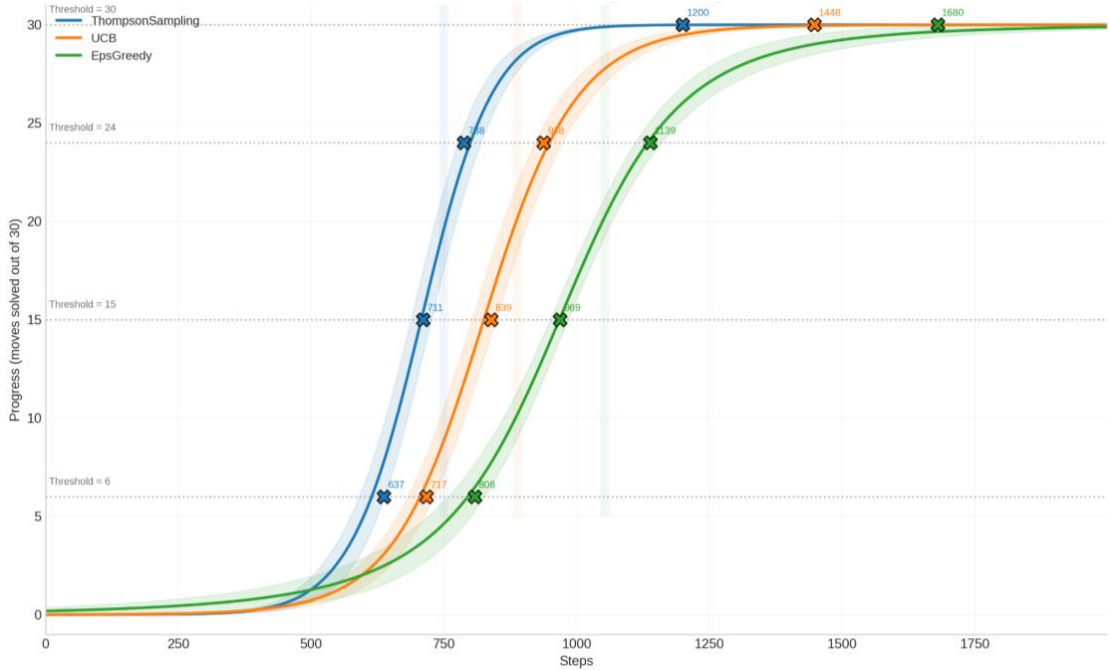

Figure 1.1: *NLL-Curriculum Leaning models* with *Run-rate trajectories showing stage-entry performance across algorithms with 95% confidence intervals.*

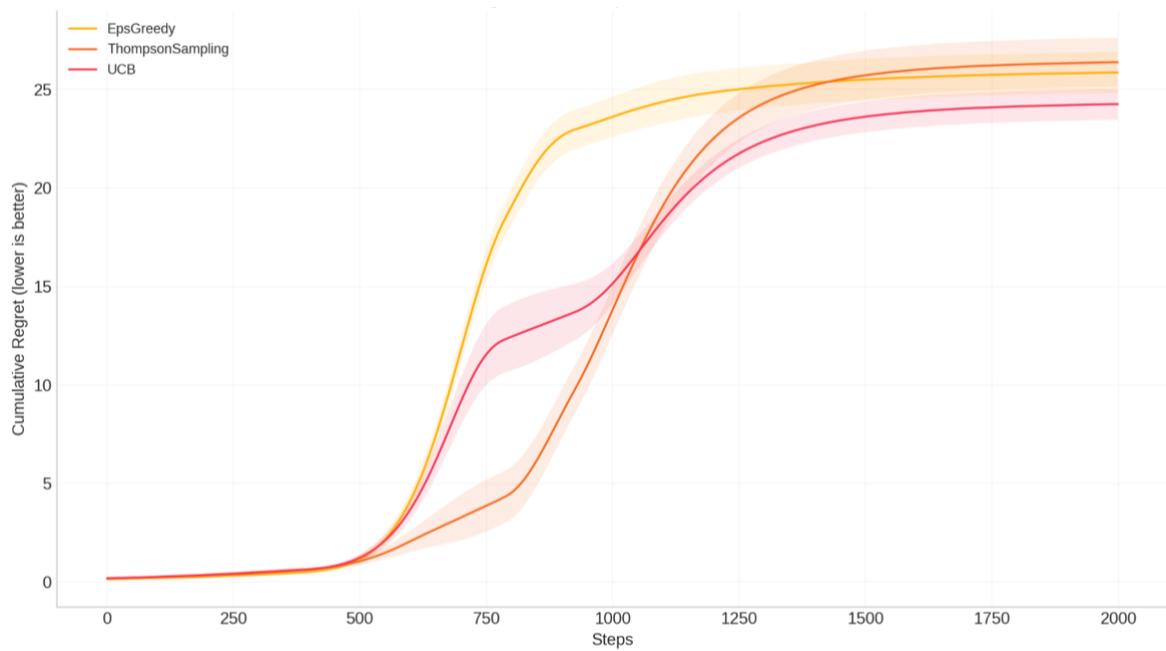

Figure 1.2: *Cumulative regret profiles comparing algorithmic efficiency over time with 95% confidence intervals.*

Algorithms with faster run-rate and lower regret directly minimize token consumption by reducing the number of model calls and wasted reasoning cycles. Thompson Sampling achieves the lowest cost because it reaches competence earliest and makes the fewest mistakes during learning.

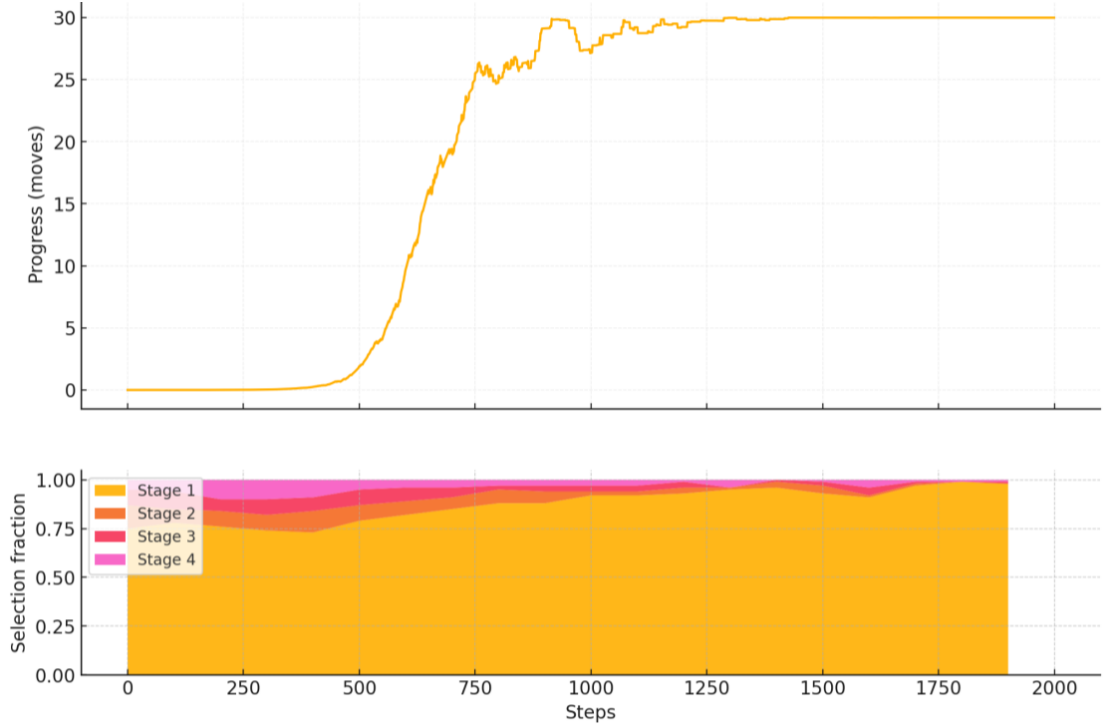

Figure 1.3: *Run-rate progression and corresponding curriculum-stage selection patterns over time.*

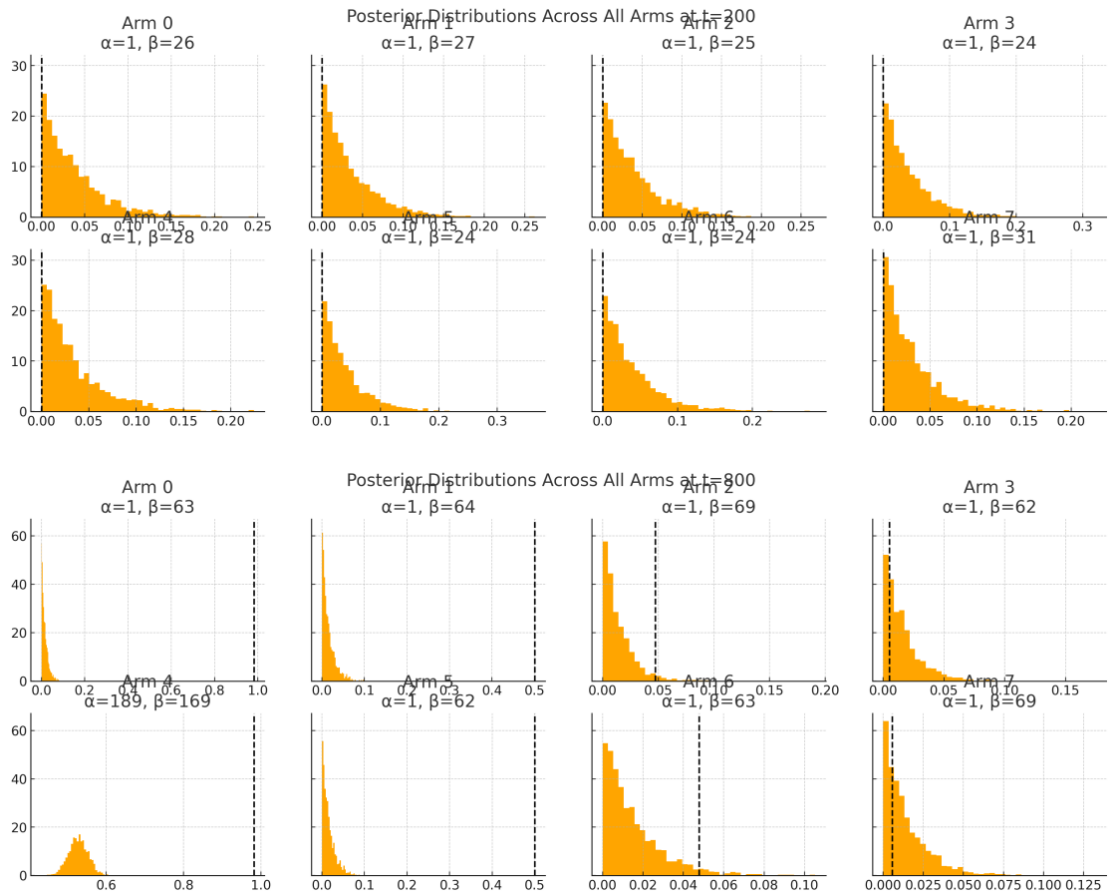

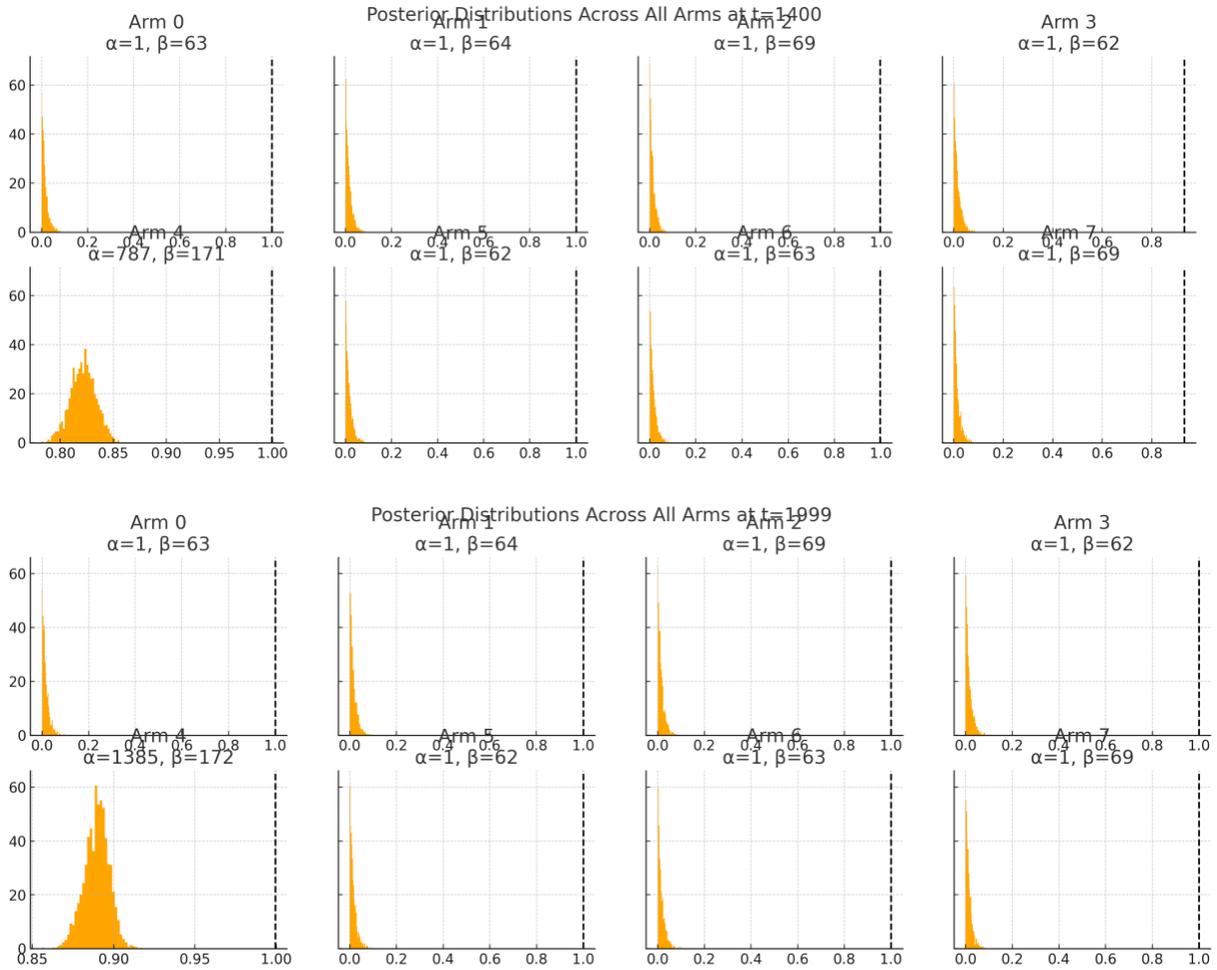

Figure 1.4: *Posterior distributions for all arms at t = 200, t = 800, t = 1400 and t = 1999*

Across time, the posterior distributions become increasingly concentrated, indicating rising confidence; however, they do not necessarily converge to the true underlying arm mean. This effect arises because Thompson Sampling prioritizes reward maximization over unbiased estimation, leading to posterior lock-in when suboptimal arms receive limited exploration.

Arm 4 received by far the most selections, accumulated the largest α count, and has the highest posterior mean (≈0.88–0.90), i.e. the posterior is both concentrated and centered near a high reward probability. This combination high selection frequency, large α, and large posterior mean identifies it as the empirical optimum under Thompson Sampling.

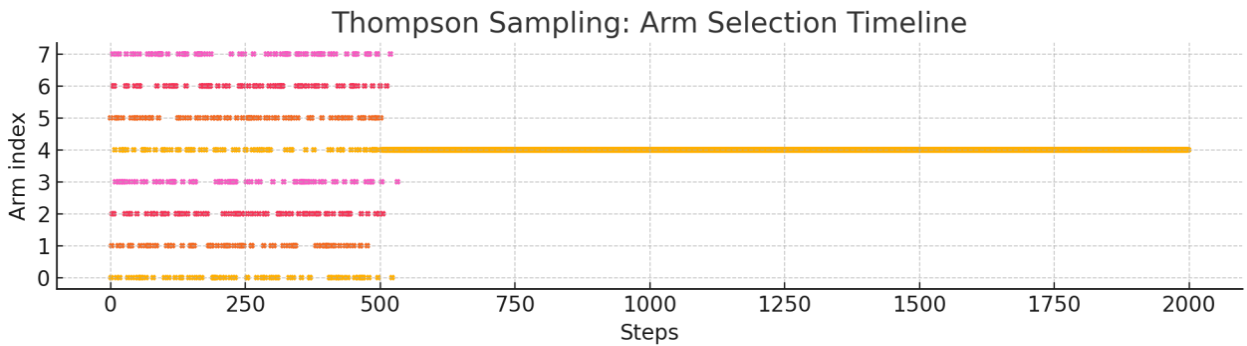

Figure 1.5: *Timeline of arm selections under Thompson Sampling, showing early exploration followed by sustained exploitation of the highest-reward arm.*

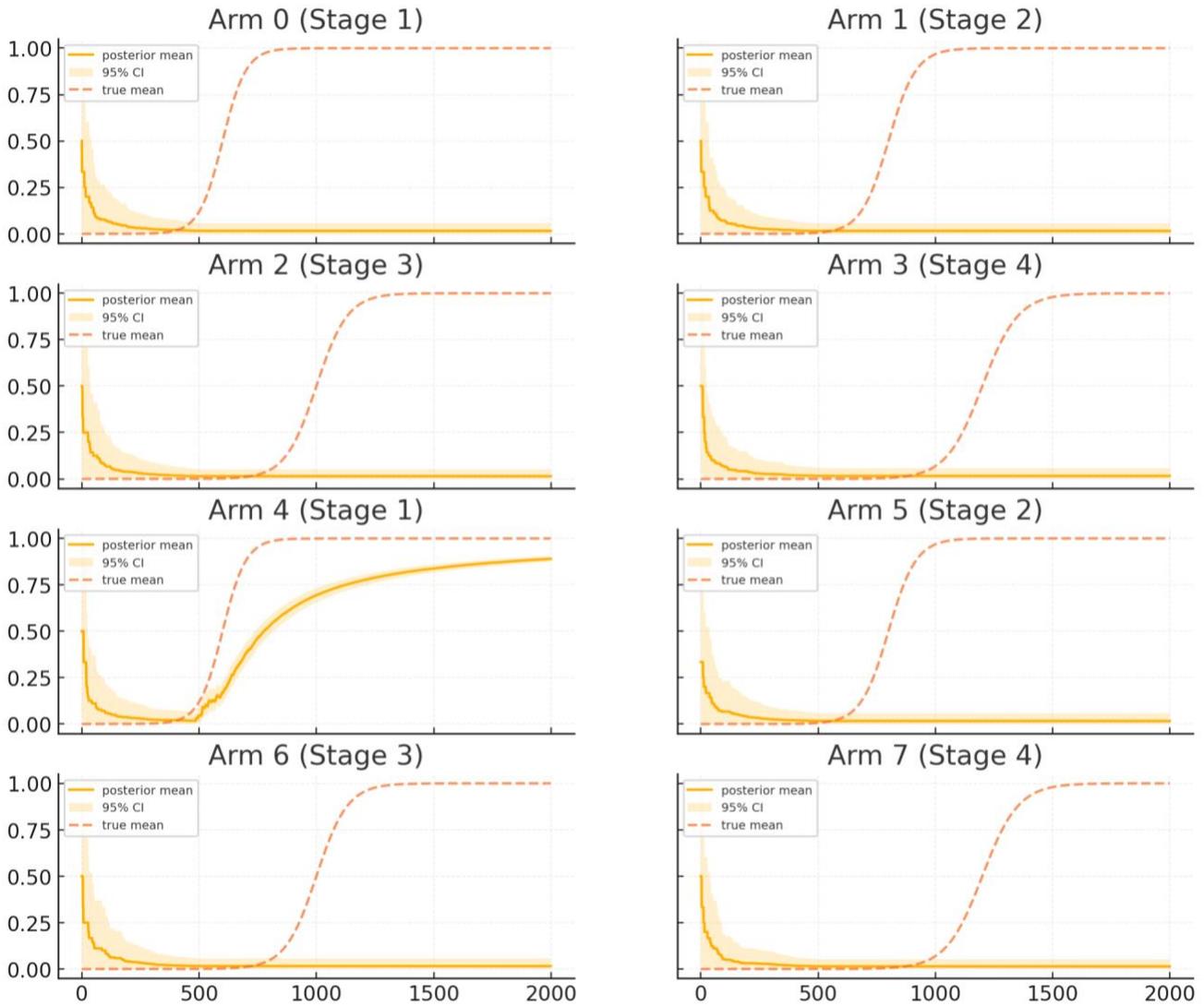

Figure 1.6 : *Posterior mean trajectories and 95% credible intervals for all eight arms under Thompson Sampling.*

Arm 4 is the optimal (highest-reward) arm in this simulation. This is consistent with the behaviour of ThompsonSampling, it concentrates samples on the arm with the highest expected reward, causing its posterior to rise while all others fall.

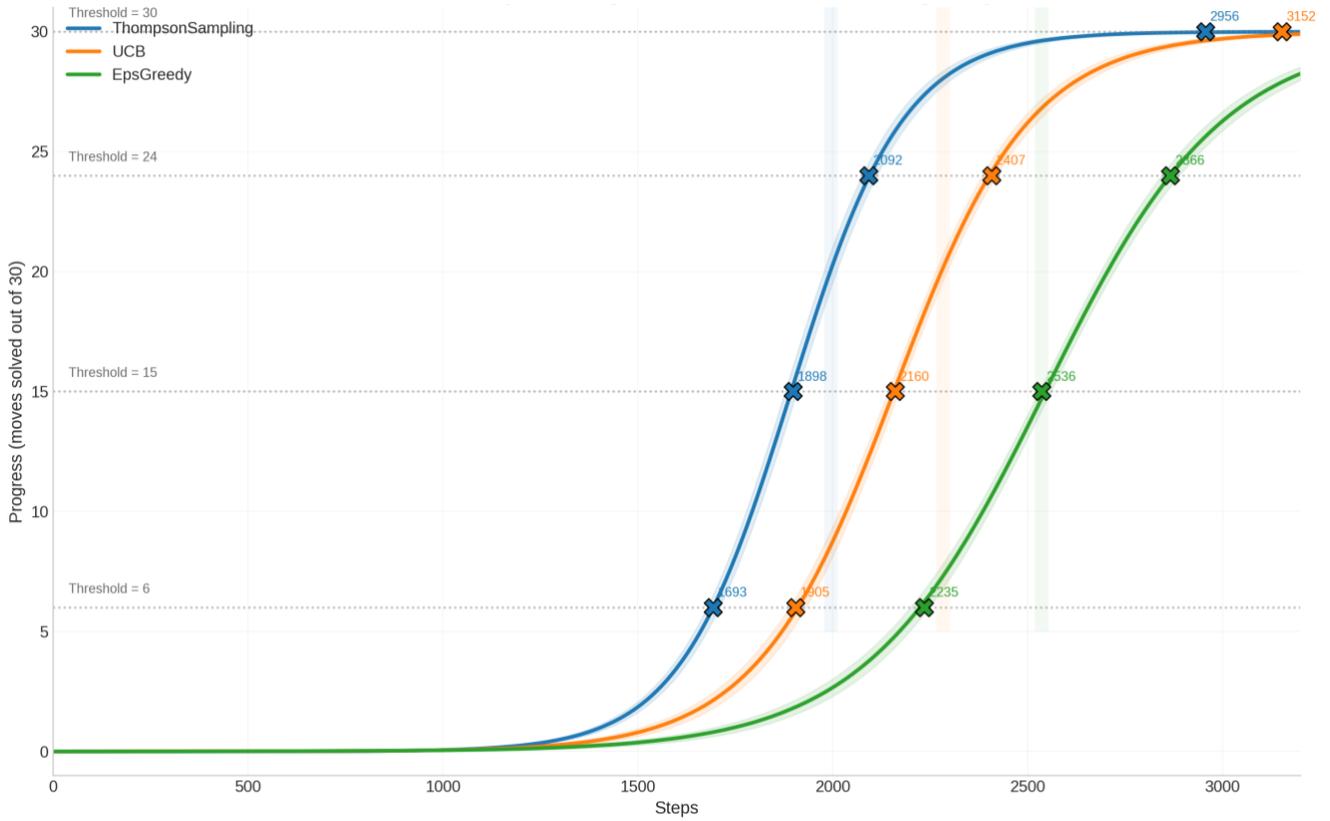

*Figure 1.7 : Only Curriculum Leaning models with Run-rate trajectories showing stage-entry performance across algorithms with 95% confidence intervals.*

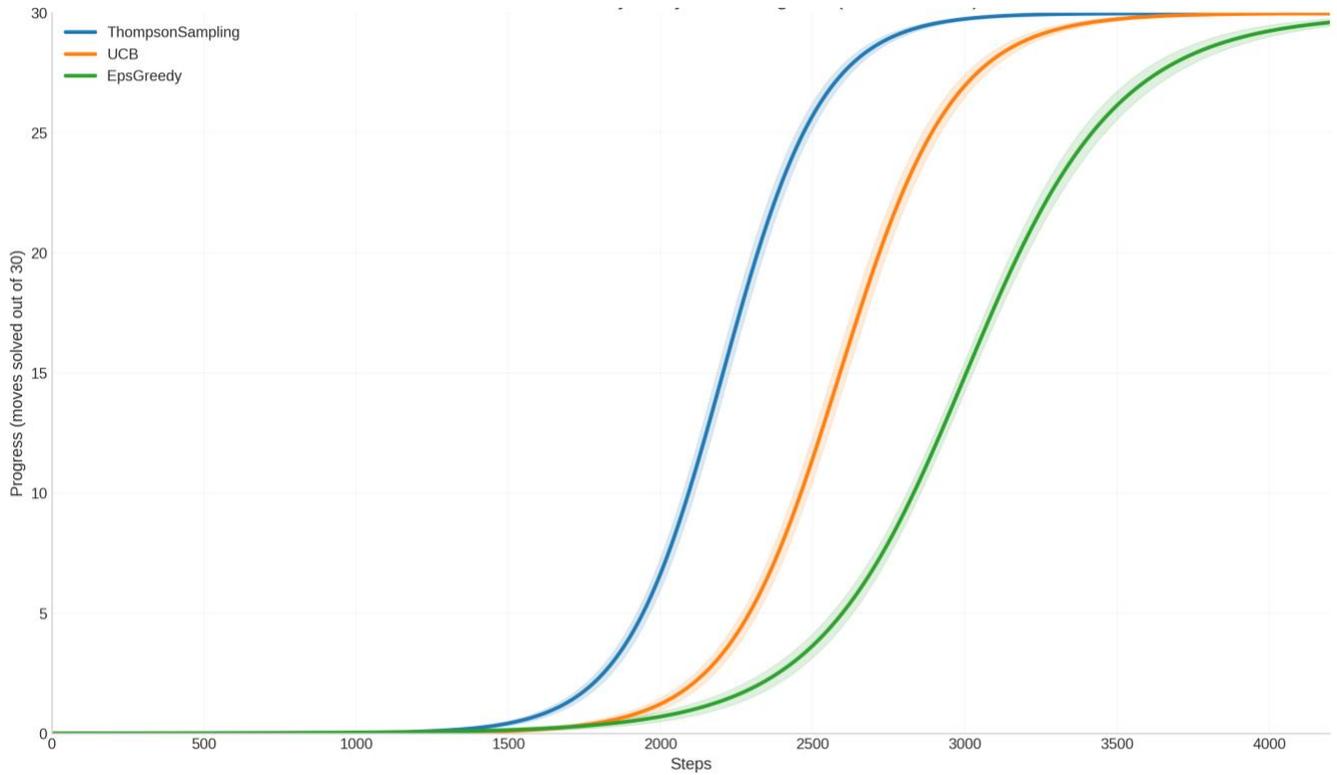

*Figure 1.8 : Base RL( no NLL no Curriculum ) models* with *Run-rate trajectories showing stage-entry performance across algorithms with 95% confidence intervals.*

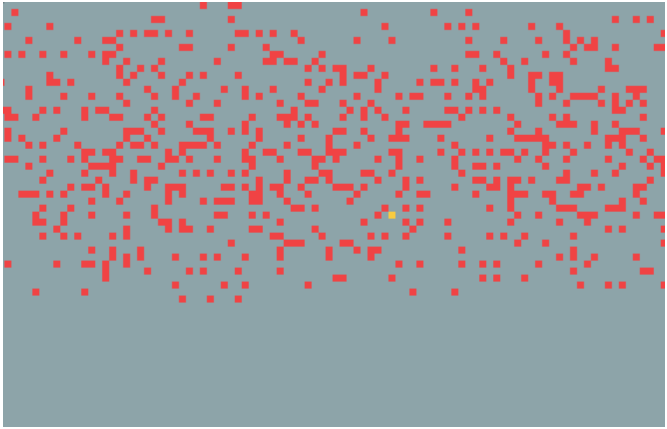
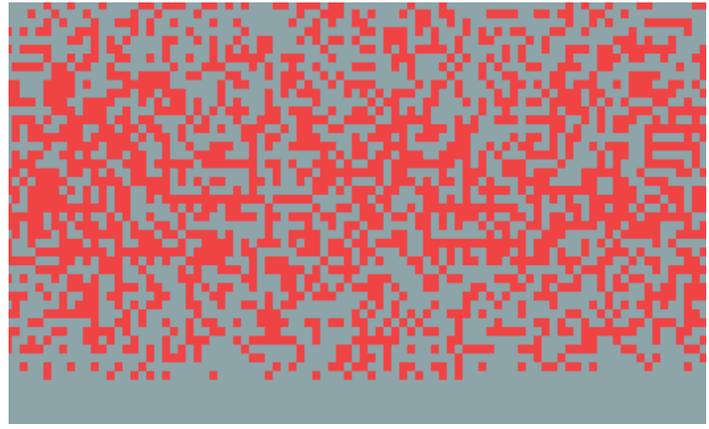

(a)                                                                       (b)

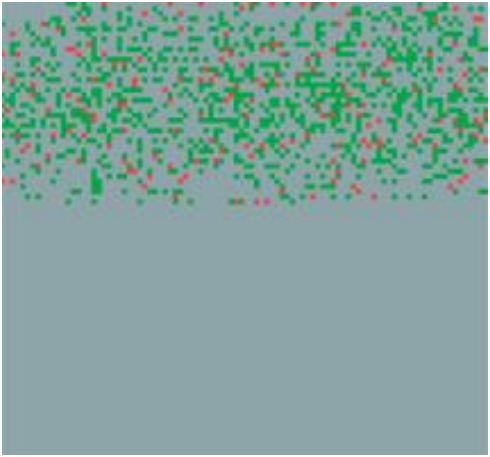
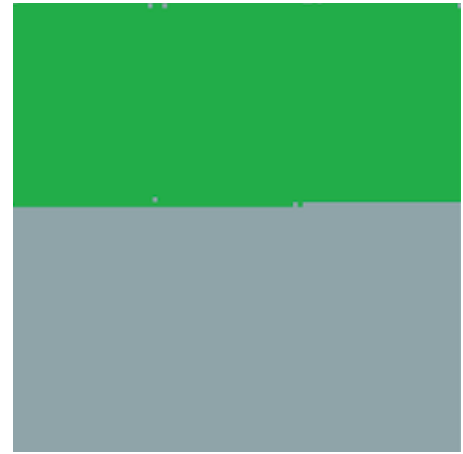

(c)                                                                       (d)

Figure 1.9:
*(a) Traditional machine-learning–based pixel agents fail to perform the micro-level task.*
*(b) Deep-learning pixel agents also fail to perform the micro-level task.*
*(c) Reinforcement-learning pixel agents without curriculum learning show limited success, with most agents still failing.*
*(d) Pixel agents trained with both curriculum learning and reinforcement learning achieve the highest success rate.*

## 6 Discussion

Our new research draws a clear commonality between the Tower of Hanoi problem and modern automotive factory AI workflows. Both systems require strict ordering, dependency management, and coordinated progression toward a final goal, which is fundamentally a long-horizon task. In Hanoi, a disk can move only once prerequisite moves are completed; similarly, factory robots must finish specific operations before others can begin, reflecting identical precedence constraints. Completion occurs only when all required actions by all agents are finished, mirroring the full-stack completion condition in Hanoi. This shared structure highlights how complex industrial processes naturally resemble hierarchical, dependency-driven problem-solving frameworks found in classical planning tasks.

Building on this analogy, the transition from symbolic puzzles to real industrial operations becomes seamless when the control policy is implemented through LLM and RL driven multi-agent systems. If the entire Tower of Hanoi sequence can be solved through a coordinated hierarchy of language-model agents, the same methodology can be transferred to robotic manufacturing systems, where each robot is governed by a lightweight Small Language Model (SLM). In this setting, our framework can be refactored so that each autonomous agent follows learned sequencing rules and executes long-horizon policies analogous to the pixel-agent behaviors in Hanoi, but grounded in physical factory tasks.

This connection also clarifies the importance of algorithmic efficiency in long-horizon environments. Algorithms with faster run-rate and lower cumulative regret directly minimize token consumption by reducing the number of model calls, incorrect rollouts, and wasted reasoning cycles. In practical deployments, every mistake or misaligned decision triggers additional inference steps, each of which incurs computational and monetary cost, before the system can recover and proceed. Thompson Sampling achieves the lowest cost profile because it reaches competence early and commits fewer errors during learning, thereby limiting exploration of suboptimal behaviors and preventing unnecessary deliberation. When scaled to factory settings, this efficiency translates to reduced token usage, lower operational overhead, and faster convergence to stable robotic workflows.

Together, these insights demonstrate that long-horizon industrial automation and classical hierarchical puzzles share a structural foundation that can be exploited by modern agentic architectures. Our methodology unifies them through a curriculum-driven, multi-agent LLM system designed to enforce ordering constraints, optimize learning efficiency, and reduce inference cost at scale.

# 7 Conclusion

The system demonstrates that long-horizon reasoning can be made tractable by combining task decomposition, curriculum learning, and probabilistic calibration within a large multi-agent architecture. By distributing the problem across thousands of micro-agents and grounding each step of the puzzle in spatial structure, the system transforms a sequential reasoning challenge into a scalable parallel process. The spatial curriculum ensures that agents encounter progressively harder regions only after demonstrating mastery in easier ones, while NLL-based calibration prevents premature advancement by accounting for both correctness and confidence. Selective escalation to a large LLM oracle provides global reasoning only when necessary, allowing the architecture to remain computationally efficient despite its scale. Together, these components form a cohesive framework that stabilizes long-horizon execution, reduces error propagation, and offers a principled path toward reliable multi-agent LLM systems.

### References

[1] S. Ling, X. Wang, et al., "ELHPlan: Efficient Long-Horizon Task Planning for Multi-Agent Collaboration," arXiv preprint arXiv:2509.24230, 2025.
[2] Y. Zeng, et al., "S²-MAD: Sparsified Multi-Agent Debate for Communication-Efficient Reasoning," NAACL, 2025.
[3] J. Lin, et al., "Stop Wasting Your Tokens: A SupervisorAgent for Token-Efficient Multi-Agent Execution," arXiv preprint, 2025.
[4] X. Zhang, et al., "MAS Failure Taxonomy: Failure Modes in LLM-Based Multi-Agent Systems," arXiv preprint, 2025.
[5] T. Yang, P. Feng, Q. Guo, J. Zhang, X. Wang, Z. Mao, "AutoHMA-LLM: Efficient Task Coordination and Execution in Heterogeneous Multi-Agent Systems Using Hybrid Large Language Models," IEEE Trans. Cognitive Communications & Networking, 2025.
[6] W. Chen, J. Yuan, C. Qian, C. Yang, Z. Liu, M. Sun, "OPTIMA: Optimizing Effectiveness and Efficiency for LLM-Based Multi-Agent System," Findings of ACL (ACL Findings), 2025.
[7] L. E. Erdogan, H. Furuta, S. Kim, N. Lee, S. Moon, G. Anumanchipalli, K. Keutzer, A. Gholami, "Plan-and-Act: Improving Planning of Agents for Long-Horizon Tasks," ICML, 2025.
[8] M. Geng, S. Pateria, B. Subagdja, L. Li, X. Zhao, A. H. Tan, "L2M2: A Hierarchical Framework Integrating Large Language Model and Multi-Agent Reinforcement Learning," IJCAI, 2025.
[9] S. Nayak, et al., "LLaMAR: Long-Horizon Planning for Multi-Agent Robots in Partially Observable Environments," NeurIPS, 2024.
[10] R. Yang, F. Zhang, M. Hou, "OceanPlan: Hierarchical Planning and Replanning for Natural-Language AUV Piloting in Large-Scale Unexplored Ocean Environments," arXiv preprint arXiv:2403.15369, (IROS submission), 2024.
[11] Y. Wang, Z. Wu, J. Yao, J. Su, "TDAG: A Multi-Agent Framework based on Dynamic Task Decomposition and Agent Generation," NeurIPS/Neural Networks journal (preprint arXiv:2402.10178), 2024.
[12] A. Anicic, et al., "HiTAMP: Hierarchical Task-and-Motion Planner Integrating LLM Planning with Robotic Skills," DYA Lab / workshop report, 2024.
[13] (ICML 2024) "cMALC-D: Contextual Multi-Agent LLM-Guided Curriculum Learning with Diversity-Based Context Blending," ICML (poster/workshop), 2024/2025 (author list as in source).

[14] (ICLR 2026) "EvoCurr: Self-Evolving Curriculum with Behavior Code Generation for Complex Decision-Making," ICLR submission / OpenReview, 2026.

[15] H. Madmoun, S. Lahlou, "Communication Enables Cooperation in LLM Agents: A Comparison with Curriculum-Based Approaches," arXiv preprint / workshop, 2024 (paper referenced in survey).

[16] X. Wu, Y. Tian, Y. Chen, P. Ye, et al., "CurriculumPT: LLM-Based Multi-Agent Autonomous Penetration Testing with Curriculum-Guided Task Scheduling," Applied Sciences (MDPI), 2023.

[17] G. Tzannetos, P. Kamalaruban, A. Singla, "Curriculum Design for Trajectory-Constrained Agents: Compressing Chain-of-Thought Tokens in LLMs," arXiv preprint arXiv:2511.02690, 2025.

[18] X. Chen, J. Lu, M. Kim, D. Zhang, J. Tang, A. Piché, N. Gontier, Y. Bengio, E. Kamalloo, "Self-Evolving Curriculum for LLM Reasoning," arXiv preprint arXiv:2505.14970 / OpenReview, 2025.

[19] E. Meyerson, G. Paolo, R. Dailey, H. Shahrzad, O. Francon, C. F. Hayes, X. Qiu, B. Hodjat, R. Miikkulainen, "Solving a Million-Step LLM Task with Zero Errors (MAKER)," arXiv preprint arXiv:2511.09030, 2025.

[20] E. Meyerson and X. Qiu, "Position: Scaling LLM Agents Requires Asymptotic Analysis with LLM Primitives," ICML position/poster (arXiv preprint), Feb. 4, 2025.

[21] T. Kwa, B. West, J. Becker, A. Deng, K. Garcia, M. Hasin, S. Jawhar, M. Kinniment, N. Rush, S. Von Arx, R. Bloom, T. Broadley, H. Du, B. Goodrich, N. Jurkovic, L. J. Miles, S. Nix, T. Lin, N. Parikh, D. Rein, L. J. K. Sato, H. Wijk, D. M. Ziegler, E. Barnes and others, "Measuring AI Ability to Complete Long Tasks," arXiv preprint, Mar. 18, 2025.

[22] R. Schaeffer, B. Miranda, and S. Koyejo, "Are Emergent Abilities of Large Language Models a Mirage?," NeurIPS Workshop / arXiv, Apr. 28, 2023.

[23] N. Dziri, X. Lu, M. Sclar, X. L. Li, L. Jiang, B. Y. Lin, P. West, C. Bhagavatula, R. Le Bras, J. D. Hwang, S. Sanyal, S. Welleck, X. Ren, A. Ettinger, Z. Harchaoui, Y. Choi, "Faith and Fate: Limits of Transformers on Compositionality," NeurIPS 2023 (Spotlight). a

[24] S. Geng, H. Cooper, M. Moskal, S. Jenkins, J. Berman, N. Ranchin, R. West, E. Horvitz, H. Nori, "Generating Structured Outputs from Language Models: Benchmark and Studies" (JSONSchemaBench), ICML (arXiv preprint), Jan. 18, 2025.

[24] OpenAI, "Introducing Structured Outputs in the API," OpenAI blog / API docs, Aug. 6, 2024. OpenAI+1

[25] Parshin Shojaee, Iman Mirzadeh, Keivan Alizadeh, Maxwell Horton, Samy Bengio, and Mehrdad Farajtabar. *The illusion of thinking: Understanding the strengths and limitations of reasoning models via the lens of problem complexity. arXiv preprint arXiv:2506.06941, 2025.*

[26] XuezhiWang, JasonWei, Dale Schuurmans, Quoc Le, Ed Chi, Sharan Narang, Aakanksha Chowdhery, and Denny Zhou. Self-consistency improves chain of thought reasoning in language models. arXiv preprint arXiv:2203.11171, 2022.

[27] Kar, Indrajit, et al. "Toward More Robust Classifier: Negative Log-Likelihood Aware Curriculum Learning." International Conference on Computational Intelligence and Data Engineering. Singapore: Springer Nature Singapore, 2022.